\documentclass[runningheads]{llncs}
\usepackage[utf8]{inputenc} 
\usepackage[T1]{fontenc}    
\usepackage{hyperref}       
\usepackage{url}            
\usepackage{booktabs}       
\usepackage{amsfonts}       
\usepackage{nicefrac}       
\usepackage{microtype}      
\usepackage{algorithm}
\usepackage{algorithmicx}
\usepackage{algpseudocode}
\usepackage{amsmath}
\usepackage{CJK}
\usepackage{graphicx}
\usepackage{amsmath,amssymb} 
\usepackage{color}
\usepackage{subfigure}
\usepackage{booktabs}
\usepackage{float}
\begin{document}
	\pagestyle{headings}
	\mainmatter
	\def\ECCVSubNumber{???}  
	
	\title{VCE: \ Variational Convertor-Encoder for One-Shot Generalization} 

	\titlerunning{VCE: \ Variational Convertor-Encoder for One-Shot Generalization}
	%
	\author{Chengshuai Li\inst{1} \and
		Shuai Han\inst{2} \and
		Jianping Xing\inst{3} }
	\authorrunning{Li, Han. et al.}
	%
	\institute{School of Microelectronics, Shandong University, Jinan, China \email{201732125@mail.sdu.edu.cn, 201932243@mail.sdu.edu.cn, xingjp@sdu.edu.cn}}
	\maketitle
	
	\begin{abstract}
		Variational Convertor-Encoder (VCE) converts an image to various styles; we present this novel architecture for the problem of one-shot generalization and its transfer to new tasks not seen before without additional training. We also improve the performance of variational auto-encoder (VAE) to filter those blurred points using a novel algorithm proposed by us, namely large margin VAE (LMVAE). Two samples with the same property are input to the encoder, and then a convertor is required to processes one of them from the noisy outputs of the encoder; finally, the noise represents a variety of transformation rules and is used to convert new images. The algorithm that combines and improves the condition variational auto-encoder (CVAE) and introspective VAE, we propose this new framework aim to transform graphics instead of generating them; it is used for the one-shot generative process. No sequential inference algorithmic is needed in training. Compared to recent Omniglot datasets, the results show that our model produces more realistic and diverse images.
		
		\keywords{Variational Convertor-Encoder, one-shot generalization}
	\end{abstract}
	\section{Introduction}
\begin{figure}[H]
		\centering
		\includegraphics[width=1.0\textwidth]{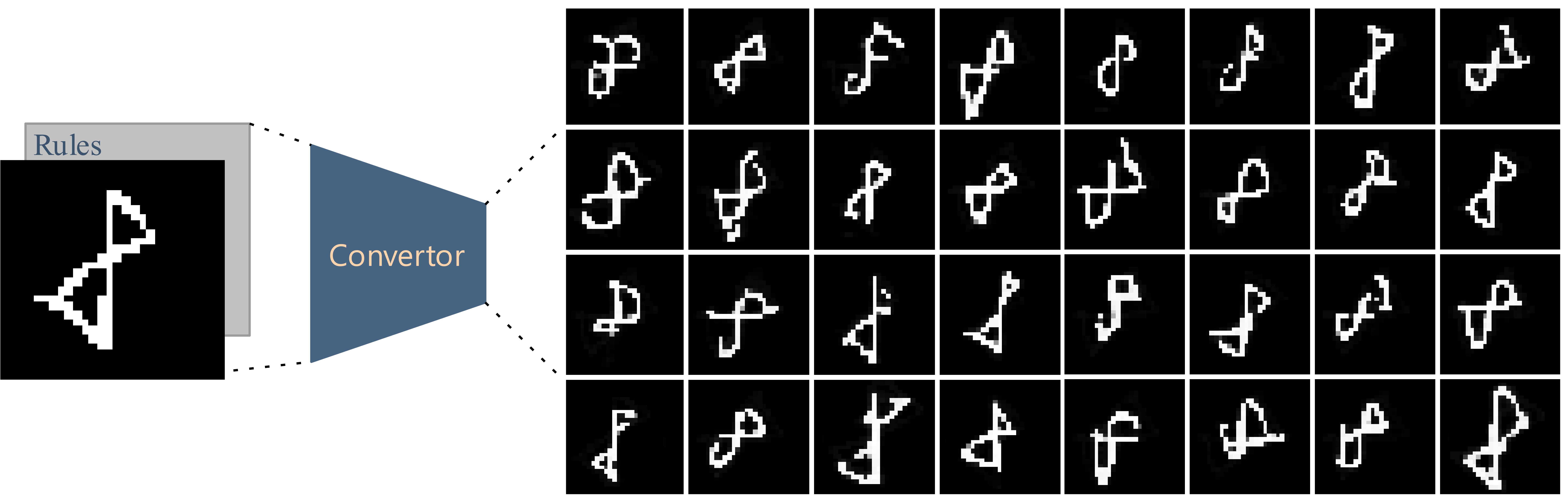}
		\caption{\textbf{The rules are sampled from the centred isotropic multivariate Gaussian, and the converter transforms the image accordingly.}}
			\label{figure1}
		\end{figure}

	The ability of humans can create images of various styles is often based on a prototype sample that they have seen before. This capability is essentially a transformation method for new models built on the experience of historical learning. We propose Variational Convertor-Encoder (VCE), as shown in Figure \ref{figure1}, the convertor can implement various styles transfer for one new image.
	
	In recent years, the deep generative model is a significant area of artificial intelligence, in that humans often have similar abilities concerning understanding and manipulating new object. Methods are used to generate images different from data sets, such as variational autoencoders (VAEs) [1, 2], and generative adversarial networks (GANs) [3], the productive process of such models is defined by a composition of conditional distributions using deep neural networks which form a hierarchy of latent and abundant of data points.
	
	Nevertheless, humans are often able to adapt to new image processing tasks on one or a small number of labelled samples [4], many techniques [5, 6, 7, 8] have been a great success in the field of image generation and rely on big datasets. However, overfitting will happen in low data regimes, even if using data augmentation and regularization techniques. On the other hand, it is more practical to generate diversified samples from a few images; this ability should be built on historical learning experience so that one-shot generalization must be adapted to accommodate new classes not seen in training.
	Some models solve these issues by conditioning, Rezende et al. [9] was able to produce images which supported conditioning on new samples and sequential generative model was used in training, Bartunov [10] was suitable for fast learning in the few-shot setting; they use matching networks share similarity concepts with new samples. [11] Uses GAN meta-trained method with Reptile [12], it is essentially a pre-training method for learning a parameter initialization that can be fine-tuned quickly on a new task.
	
	Compared with the approaches mentioned above, our main contributions are:
	\begin{itemize}
		\item We develop a novel method for one-shot image generation. It can transfer to new tasks not seen before without additional training. 	
		\item Our framework does not require an extensive sequential inference algorithmic like LSTM [13] for training.
		\item Our VCE architecture reaches a better balance between VAE and GAN; its results are precise and diverse using the LMVAE algorithm proposed by us.
	\end{itemize}
	
	We present Variational Convertor-Encoder (VCE); a purpose is considered in this structure because we aim to transform samples instead of generating them. To filter the blurred images produced by Original VAE, we propose our LMVAE algorithm, who is based on introspective VAE [15]; we improved it and offer our more straightforward method; it produces sharp and diverse results, the encoder serves as the discriminator, then blurry points are assigned to a low probability, we then propose a style regularization term for our VCE, it restrains the diversity of results but makes them more realistic.
	
	The rest of the paper is structured as follows. Section \ref{section2} describes background. The detail of our methods is introduced in Section \ref{section3}. We then report our experiments in Section \ref{section4}. Finally, Section \ref{section5} discusses our conclusions.
	
	\section{Background}
	\label{section2}	
	The process of learning can be regarded as a probabilistic transformation model which can be expressed as a conditional probability distribution ${q_\theta }(z\left| {x,} \right.q)$, who maps the $x$ and $q$ parameterized by $\theta $ to the latent variables $z$, variational convertor is conducted to indicate conditional conversion distributions of the form.
	\begin{equation}
		{p_\phi }(\left. x \right|q){\rm{ = }}\int {p(z){p_\phi }(\left. x \right|z,q)dz},
	\end{equation}
	
	where $p(z)$ is $N(0,I)$ and ${p_\phi }(\left. x \right|z,q)$ reconstructs samples $x$ given $z$ and the condition $q$.
	VAE maximizes the marginal likelihood $\log {p_\phi }(\left. x \right|q)$ by approximating it with the variational lower bound [16, 17]:
	\begin{equation}
		\log {p_\phi }(\left. x \right|q) \ge {E_{{q_\theta }(z\left| {x,} \right.q)}}\log {p_\phi }(\left. x \right|z,q) - KL(\left. {{q_\theta }(z\left| {x,} \right.q)} \right\|p(z)).
	\end{equation}

	Reptile is currently a widely used approach for few-shot learning, it learns an initialization that is good for fine-tuning, and the gradient updates correspond to
	\begin{equation}
		{g_{k > 1}} = {E_\tau }[\phi  - U_\tau ^k(\phi )]/\alpha,
	\end{equation}

	where $U_\tau ^k(\phi )$ denotes $k$ steps of SGD and $\alpha $ is the stepsize, Reptile maximizes within-task generalization by maximizes the inner product between the gradients on different batches from the same task [12].
	
	\section{Methods}
	\label{section3}
	Our approach is used for one-shot generalization without additional training and adapted to accommodate new classes not seen in training. This method consists of three components, which are described in the following subsections. First, we describe how the primary process is defined, and we explore recent advances in condition variational auto-encoder (CVAE) and IntroVAE, as discussed in Section 1. We show how we improve them for one-shot generalization, and we extend them to our VCE models. We further employ our training strategy for the problem of one-shot generalization.
	
	\subsection{Basic method}
	
	The prior $p(z)$ in our model is the same as the original VAEs [1], that is the $D$ dimension centered isotropic multivariate Gaussian $N(0,I)$, then the input $z$ of convertor is obtained by $z = \mu  + \sigma  \odot \varepsilon $, where $\mu $ and $\sigma $ are from the posterior ${p_\phi }(\left. x \right|z,q)$ and $\varepsilon  \sim N(0,I)$, in this setup, the error of KL-divergence can be computed as below:	
	\begin{equation}
		{L_{KL}}{\rm{ = }}KL(\left. {{q_\theta }(z\left| {x,} \right.q)} \right\|p(z)) = \frac{1}{2}\sum\limits_{i = 1}^D {(\mu _i^2 + \sigma _i^2 - \log (\sigma _i^2) - 1)},
	\end{equation}

	we select the multivariate Bernoulli as the ${p_\phi }(\left. x \right|q)$ computed from the latent variables $z$, the conversion error is defined as:	
	\begin{equation}
		{L_{CON}} =  - \log {p_\phi }(\left. x \right|z,q) = \sum\limits_{i = 1}^M {[ - {x_i}\log {y_i} - } (1 - {x_i})\log (1 - {y_i})],
	\end{equation}
	
	where ${y_i}$ is converted sample and $M$ is the dimension of the data ${x_i}$.
	
	GAN usually generates highly distinct images, but it is hard to stabilize in a new training assignment, and it lacks sampling diversity, which is a more profound challenge in our one-shot training tasks. In contrast, VAE is easy to train, and most of the images it generates are blurred; one of conjecture is that VAE cannot ensure blurry points are assigned to a low probability [18], then add a discriminator to add an adversarial constraint into data or latent variables [6, 7, 8] is a direct method of filtering blurred images, unlike these hybrid models of VAE and GAN, the encoder of Introspective VAE itself serves as a discriminator.
	
	Similar to GAN, the coefficients of encoder and convertor are updated alternately, especially the encoder distinguishes the real images $x$ from converted images ${x_c}$ and sampled images ${x_s}$ in training, the loss functions for encoder and convertor are computed as:
	
	\begin{equation}
		{L_E} = {L_{KL}}(z) + \alpha \sum\limits_{j = c,s} {\max [0,m - {L_{KL}}({z_j})] + \beta {L_{CON}}(x,{x_c})},
	\end{equation}
	
	\begin{equation}
		{L_C} = \alpha \sum\limits_{j = c,s} {{L_{KL}}({z_j}) + \beta {L_{CON}}(x,{x_c})},
	\end{equation}

	where $m$ is margin and ${z_s}$ samples from prior $N \sim (0,I)$, $\alpha $ and $\beta $ are hyper-parameters denote the importance of each item.
	
	\subsection{Our full model}
	
	Using the basic introspective VAE described in the previous section, the training is challenging to hold no matter how we adjust those hyper-parameters. Our experiments eventually produce two kinds of terrible results; there are blurred images and sharp images but lacking diversity. We think they designed a model closer to GAN using VAE, but we need it is more like VAE and just filtering out a few blurred images in training, then a simpler model is proposed by us:
	\begin{equation}
		{L_C} = {L_{KL}}(z) + \lambda {L_{KL}}({z_c}) + {L_{CON}}(x,{x_c}),
	\end{equation}

	\begin{equation}
		{L_E} = {L_{KL}}(z) + \lambda \max [0,m - {L_{KL}}({z_c})] + {L_{CON}}(x,{x_c}).
	\end{equation}

	We named it a large margin VAE (LMVAE) loss function, and our LMVAE achieves a balance between VAE and GAN; the only difference between our model and original VAE is that we filter out blurred converted images by this confrontation.
	
	Our model can behold just by adjusting the margin appropriately. Theorem 1 is still valid as proved by [15], it indicates that when the model approximately converged, ${L_{KL}}(z), \le m$ and ${L_{KL}}({z_c})$ is close to $m$, diversified images with higher quality are generated along with the training process.
	
	{\bfseries Theorem 1.} $({E^ * },{C^ * })$ forms a saddle point of the above system if and only if ${p_{{C^ * }}} = {p_{data}}$ and ${E^ * }(x) = \gamma $, where $\gamma  \in [0,m]$.
	
	The latent variables ${z_c}$ are the rules of style conversion between samples of the same class; however, imposing an overly complex transformation on a new image usually results in over-distortion. We propose the style regularization term to solve this problem:
	
	\begin{equation}
		{L_{REG}} = \sum\limits_{i = 1}^M {[ - {q_i}\log {y_i} - } (1 - {q_i})\log (1 - {y_i})],
	\end{equation}

\begin{figure}[H]
	\centering
	\includegraphics[width=1.0\textwidth]{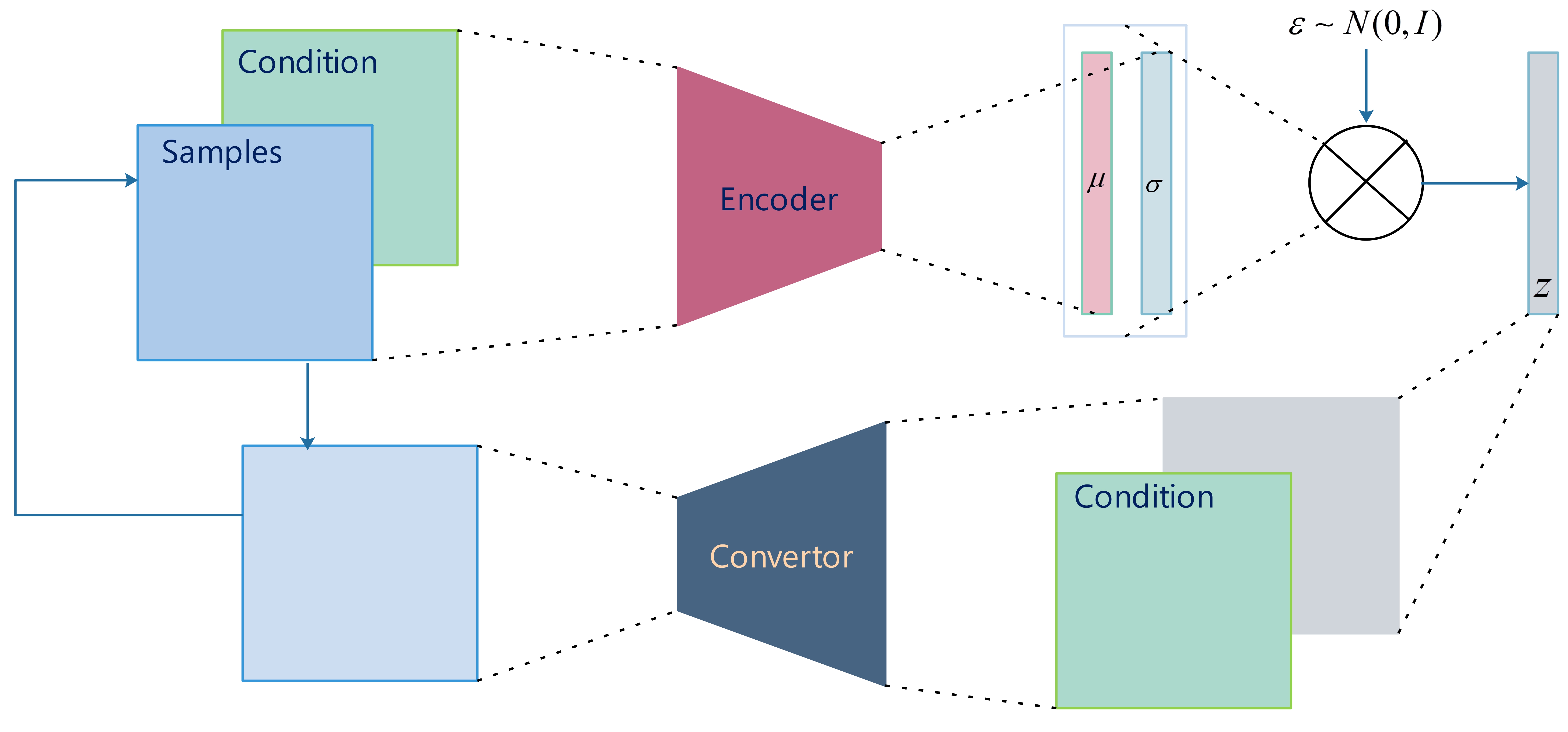}
	\caption{\textbf{A VCE with encoder and convertor, the encoder maps differences to latent variables, and the condition is converted by the noise sampled in latent $z$.}
		\label{figure2}}
	\end{figure}

	Where ${y_i}$ is converted sample and ${q_i}$ is the data of condition sample, this ${L_{REG}}$ restrain the diversity of ${y_i}$, but it helps to generate more realistic images. Finally, our full VCE loss functions for encoder and convertor are defined as:
	
	\begin{equation}
		{L_E} = {L_{KL}}(z) + \lambda \max [0,m - {L_{KL}}({z_c})] + (1 - \sigma ){L_{CON}}(x,{x_c}) + \sigma {L_{REG}}(x,{x_c}),
	\end{equation}

	\begin{equation}
		{L_C} = {L_{KL}}(z) + \lambda {L_{KL}}({z_c}) + (1 - \sigma ){L_{CON}}(x,{x_c}) + \sigma {L_{REG}}(x,{x_c}),
	\end{equation}

	we usually set $\sigma $ to be less than 0.2. Figure \ref{figure2} present a diagram of the process, and the full algorithm is shown in algorithmic \ref{algorithm1}.
	
\begin{CJK*}{UTF8}{gkai}
	\begin{algorithm}
		\caption{Training episode loss algorithm for VCE. $N_C$ is a set of samples of one class per episode,
			$S_k$ is a set of support images each class, $Q_c$ is one condition image per class.}
		\label{algorithm1}
		\begin{algorithmic}[1] 
			\Require Training set $I = \{ {x_1},...,{x_N}\} $, encoder $Enc$, convertor $Con$
			\Ensure The reconstruction error ${L_C}$ and the KL-divergence ${L_E}$
			\State {Random selection of $N_C$ from Training set $I$}
			\State {Random selection of $Q_c$ from $N_C$, Random selection of $S_k$ from $N_C$}
			\For{$k$ in $S_k$}
			\State ${Z_k} \leftarrow Enc({S_k},{Q_c})$, ${X_k} \leftarrow Con({Z_k},{Q_c})$ 
			\State $Z_k^r \leftarrow Enc({X_k},{Q_c})$
			\State${L_E} \leftarrow {L_{KL}}({Z_k}) + \lambda \max [0,m - {L_{KL}}(Z_k^r)] + (1 - \sigma ){L_{CON}}({X_k}) + \sigma {L_{REG}}({X_k})$
			\State${\tilde \theta _E} \leftarrow {\theta _E} - \eta \frac{{\rm{1}}}{{{S_k}}}{\nabla _{{\theta _E}}}{L_E}$
			\State$Z_k^r \leftarrow Enc({X_k},{Q_c})$
			\State${L_C} \leftarrow {L_{KL}}({Z_k}) + \lambda {L_{KL}}(Z_k^r) + (1 - \sigma ){L_{CON}}({X_k}) + \sigma {L_{REG}}({X_k})$
			\State${\tilde \theta _C} \leftarrow {\theta _C} - \eta \frac{{\rm{1}}}{{{S_k}}}{\nabla _{{\theta _C}}}{L_C}$
			\EndFor
		\end{algorithmic}
	\end{algorithm}
\end{CJK*}
	\subsection{Training strategy}	
	In our VCE training strategy, two images with the same property are encoded together, and one of them is input to convertor as a condition; and note that each training batch consists of the same class of samples, and every set of inputs in a batch shares one condition sample, it ensures that each of them consists of different transformation rules between samples.
	
	Only residual structure [14] and one stride convolution layers are used in the converter so that all information, including coordinates, will be preserved to the end; we represent the noisy output of the inference model as transformation rules between images of the same class, then convertor is required to processes the condition from noisy. Moreover, we do not use stochastic gradient descent directly on our models; we pre-train VCE a few epochs in the original VAE loss function with Reptile, as shown in algorithmic \ref{algorithm2}.
	\begin{CJK*}{UTF8}{gkai}
		\begin{algorithm}
			\caption{Pre-training episode loss algorithm for VCE. $N_C$ is a set of samples of one class per episode,
				$S_k$ is a set of support images each class, $Q_c$ is one condition image per class.}
			\label{algorithm2}
			\begin{algorithmic}[1] 
				\Require Training set $I = \{ {x_1},...,{x_N}\} $, encoder $Enc$, convertor $Con$
				\Ensure The reconstruction error ${L_C}$ and the KL-divergence ${L_E}$
				\State Initialize $\phi $ , the vector of initial parameters
				\State {Random selection of $N_C$ from Training set $I$}
				\State {Random selection of $Q_c$ from $N_C$, Random selection of $S_k$ from $N_C$}
				\For{iteration = 1, 2, $ \ldots $}
				\For{$k$ in ${S_k}$}
				\State ${Z_k} \leftarrow Enc({S_k},{Q_c})$, ${X_k} \leftarrow Con({Z_k},{Q_c})$
				\State$\tilde \phi  \leftarrow \phi  - \eta \frac{{\rm{1}}}{{{S_k}}}{\nabla _\phi }({L_{CON}}({X_k}) + {L_{KL}}({Z_k}))$
				\EndFor
				\EndFor
				\State Update $\phi  \leftarrow \phi  + \alpha {\rm{(}}\tilde \phi  - \phi )$
			\end{algorithmic}
		\end{algorithm}
	\end{CJK*}
	
	Each training batch is formed by randomly selecting a support set and a condition point from the one class of training set; the encoder input consists of the condition point and one support point; this ensures that each group of information in a batch consists of different transformation rules between samples of one class. In particular, the number of our latent variables is twice the width of the image; every two sampled noises are juxtaposed with one line of conditional image and input to convertor as shown in Figure \ref{figure2}; we found that this approach yielded better results, and we conjecture that it encourages the converter to learn the translation information of pixels in two dimensions.
		
	\section{Experiments}
	\label{section4}
	In this section, we first detail the data preprocessing of training and testing datasets. Next, we show the training details, such as hyperparameters settings. Subsequently, we perform our results to evaluate the performance of VCE; we show the advantages of our LMVAE loss for the problem of one-shot generalization.
	\subsection{Data preprocessing}
	The experiments were conducted on the Omniglot dataset, it contains different handwritten characters from 50 different alphabets, each of those characters was drawn by 20 different people, we were using 30 alphabets for training and 20 for the test, as suggested by [19]. We resized the grayscale images to 28*28, and we did not augment those training data with any further preprocessing.
	\subsection{Training details}
\begin{figure}[htpb]
		\centering
		\includegraphics[width=1.0\textwidth]{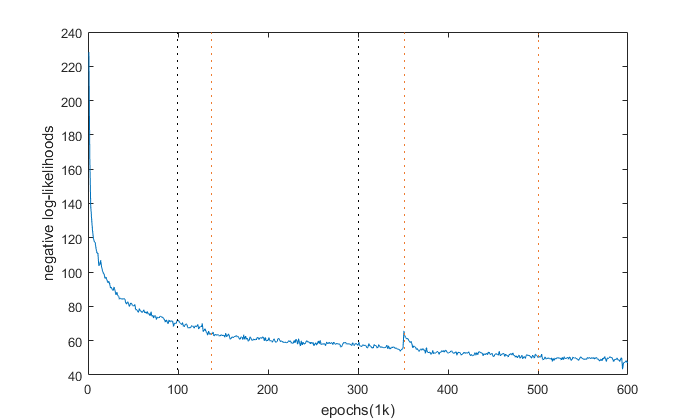}
		\caption{\textbf{The training negative log-likelihoods as a function of iteration.}
			\label{figure3}}
	\end{figure}

	Both of our encoder and convertor are composed of 8 residual blocks and one $5 \times 5$ convolution layer. All of those filters are the SAME padding with one stride, each residual block is consists of two $3 \times 3$ convolution layers, and the number of their channels is equal to 32; this kind of structure is because we aim to preserve all information of the data including coordinates in the whole process, so that our model does not need to generate a completely new image, but does some pixels transformations based on the samples. Figure \ref{figure3} shows the convergence of negative log-likelihoods on Omniglot; we trained the model for 600, 000 steps took about 0.5 days on a single NVidia GTX1070Ti GPU, and the unit of abscissa is one thousand training epochs; their values represent the mean NLL of the previous 1000 epochs, the black lines separate two different training stages as described in the next section, the orange lines show where the learning rate was halved.

	\subsection{Evaluating models}
\begin{figure}[H]
		\centering
		\includegraphics[width=1.0\textwidth]{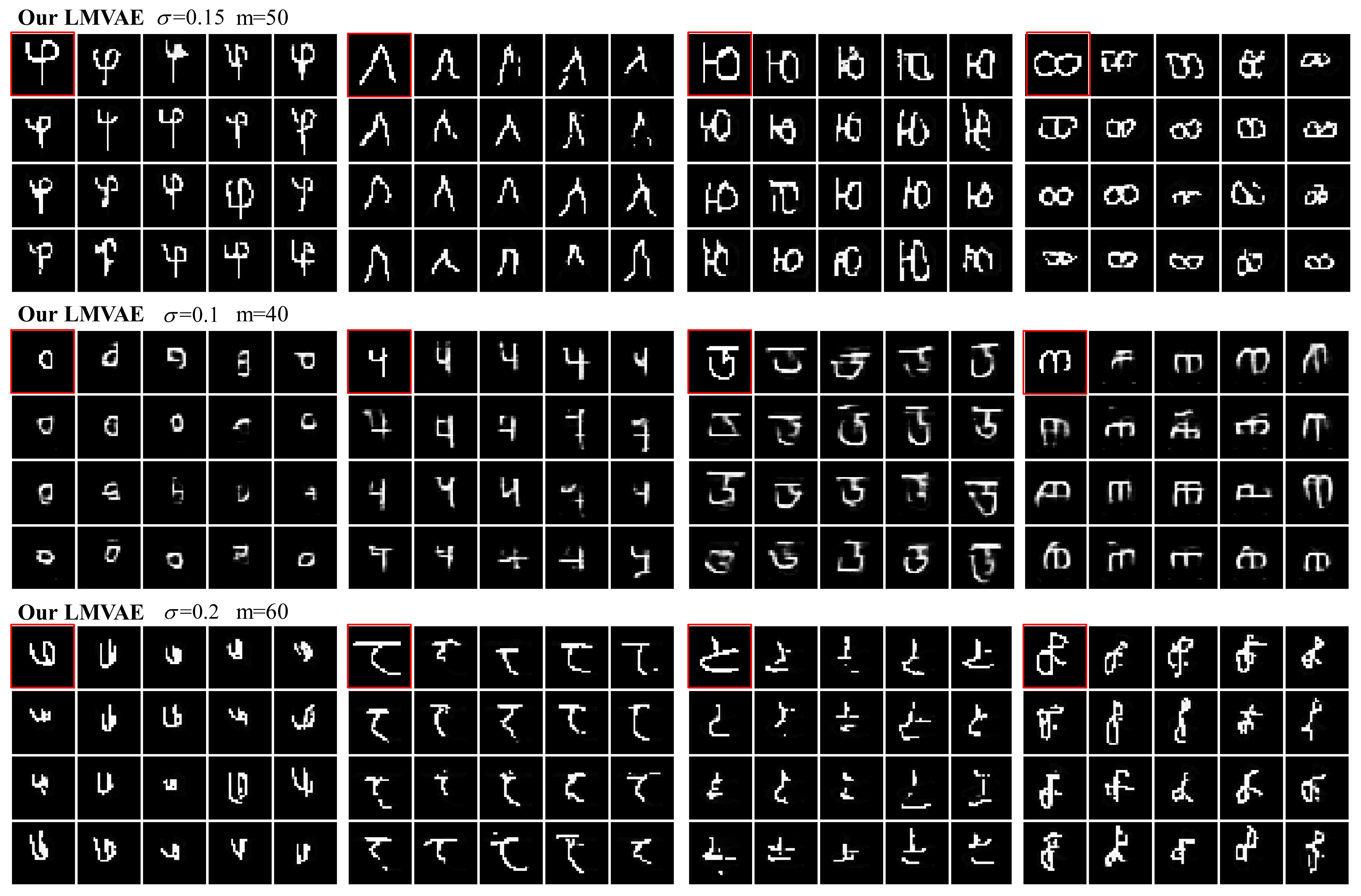}
		\caption{\textbf{The test results of different hyper-parameters on $28 \times 28$ omniglot.}}
		\label{figure4}
	\end{figure}

\begin{figure}[H]
		\centering
		\includegraphics[width=1.0\textwidth]{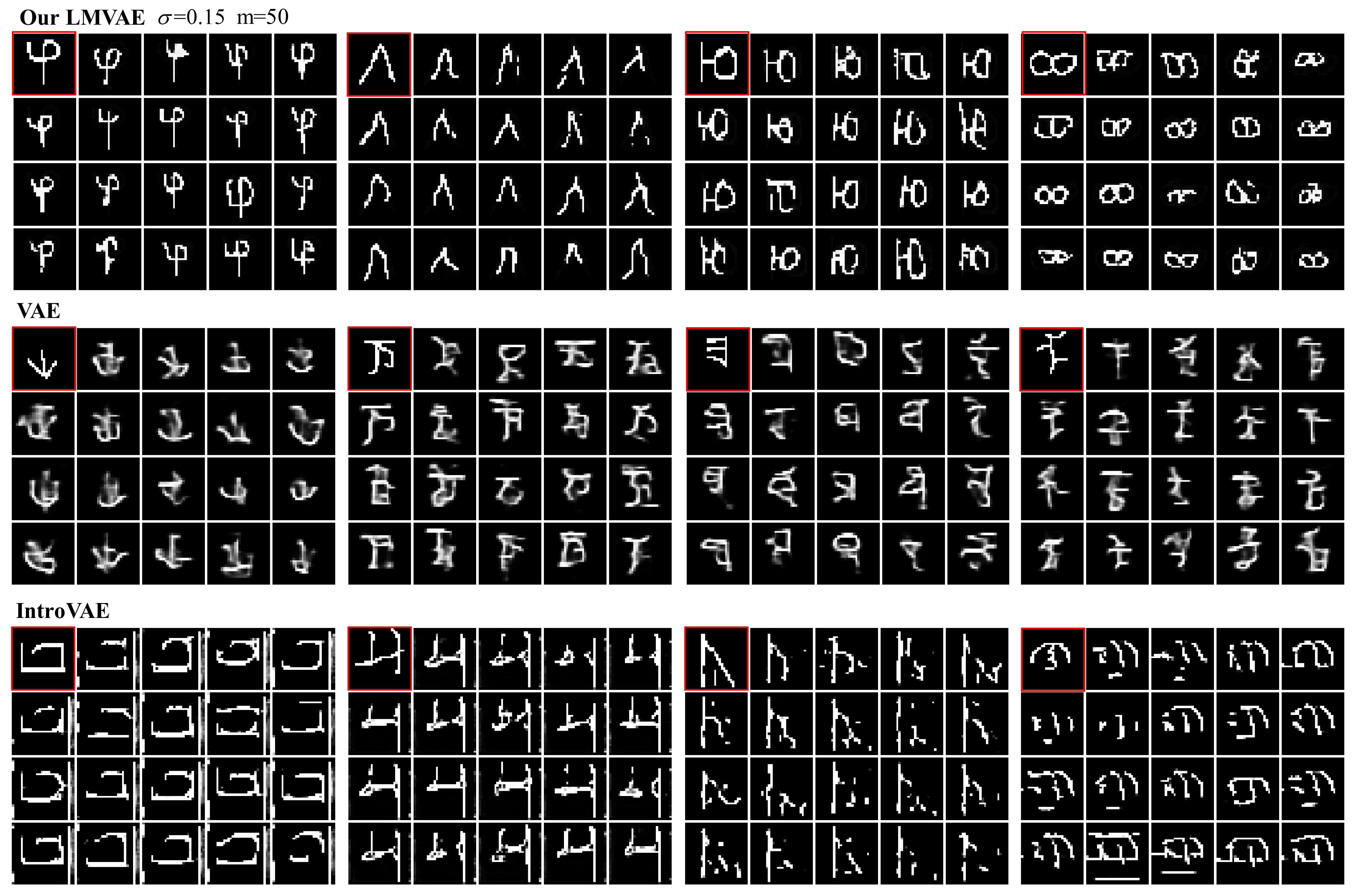}
		\caption{\textbf{The test results and comparing of different methods on $28 \times 28$ omniglot.}}
		\label{figure5}
	\end{figure}

	For VCE, we implement three different phases for the training. Initially, we pre-train our model a few epochs in the original VAE with Reptile as detailed in the algorithmic \ref{algorithm2}, and then we train VCE with usually VAE loss function until it converges; finally, we continue to improve the quality of the output images by using our LMVAE loss function as described in Section , we were using SGD with Adam optimizer [20], the key is to choose $m$ carefully, an excessive margin leads to sharp but bad results, we set the margin $m = 50$, $\sigma = 0.15$ and $\lambda = 0.2$ with a batch size of 19 and the learning rate was initialized to 0.0002. Results of different hyper-parameters are shown in Figure \ref{figure4}, and we compared our LMVAE loss function to other algorithms with the same VCE training strategy as proposed by us. The test results are shown in Figure \ref{figure5}.
	
	Compared with other methods, the results, as in our experiments, show that our model achieved sufficiently clear and diverse products. VCE converted a sample in the red box into various handwritten styles as learned in training, and it is evident that the results of VAE are too blurred and Introspective VAE output the bad test sample almost directly, and the comparison of the negative log-likelihoods for the test of Omniglot is shown in Table \ref{table1}. A model that has higher NLL in one-shot generation test does not necessarily mean that the results are worse than lower NLL models, as shown in Figure \ref{figure4}, we prevent the output of over-deformation then got more realistic results.
	\begin{table}
		\caption{The test negative log-likelihoods for the One-shot Generalization on $28 \times 28$ Omniglot}
		\label{table1}
		\centering
		\begin{tabular}{lll}
			\toprule
			Model                & Loss function           & NLL \\
			\midrule
			VAE                  &                & 106.31 \\
			Seq Gen Model [9]    &                & 95.5 \\
			GMN [10]             &                & 83.3 \\
			Our VCE              & IntroVAE       & 117.1 \\
			Our VCE              & VAE            & 68.75 \\
			Our VCE              & Our LMVAE, $\sigma  = 0.15$    & {\bf 81.39} \\
			Our VCE              & Our LMVAE, $\sigma  = 0$   & {\bf 62.8} \\
			\bottomrule
		\end{tabular}
	\end{table}
	\begin{figure}[H]
	\centering
	\includegraphics[width=1.0\textwidth]{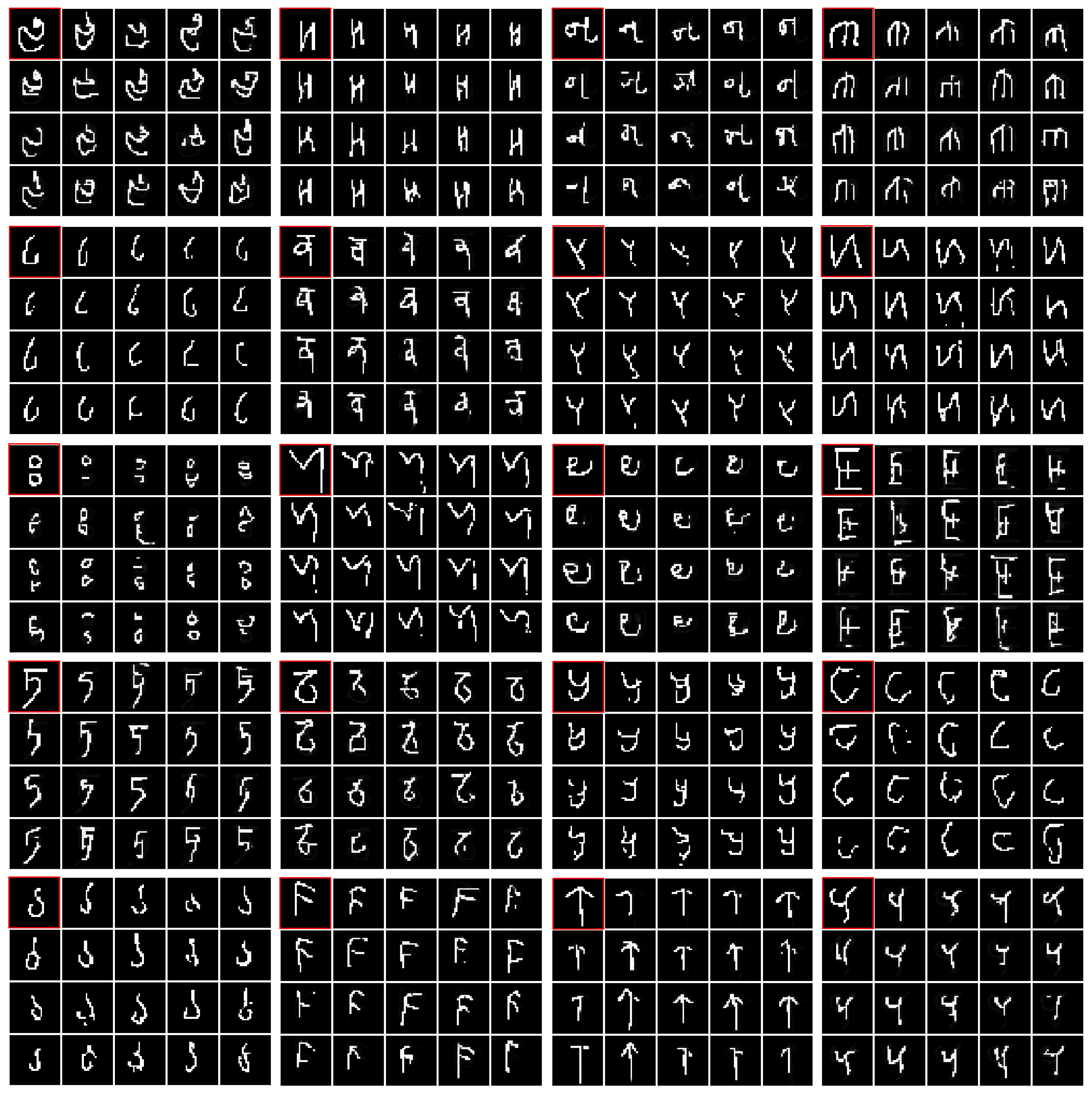}
	\caption{\textbf{Additional test results on  $28 \times 28$ omniglot.}
		\label{figure6}}
	\end{figure}
	\section{Conclusion}
	\label{section5}
	In this paper we introduced a new deep generative model, Variational Convertor-Encoder, is capable of better performance on one-shot generation tasks compared to other architectures. We explored this method to converted a new image instead of generated it, those transformation rules were learned in our VCE training strategy. Then we proposed our large margin VAE (LMVAE) loss function to improve the performance of original VAE, the most of blurred results were filtered by this algorithm, and we stabilize the diversity of the output by our style regularization term, compared to IntroVAE, our output samples are more diverse and our LMVAE is a simpler algorithm. The VCE model consists of encoder and convertor, it achieves one-shot generation without sequential inference algorithmic like LSTM and reaches a better balance between VAE and GAN, results are reached in our one-shot generation experiments. We believe that this ideas can evolve further and feel this is a challenging area which we hope to keep improving in future work.
	
	\subsubsection{Acknowledgments}
	This work is supported by China Postdoctoral Science Foundation funded project
	(2016M601152), the National Natural Science Foundation of China under grant 61603215 and
	Shandong Province independent innovation major project.
	
	\section*{References}
	\small
	[1] Kingma, Diederik P and Welling, Max. Auto-encoding variational bayes. In ICLR, 2014.
	
	[2] Rezende, Danilo Jimenez, Mohamed, Shakir, and Wierstra, Daan. Stochastic backpropagation and approximate inference in deep generative models. In ICML, pp. 1278-1286, 2014.
	
	[3] Goodfellow, Ian, Pouget-Abadie, Jean, Mirza, Mehdi, Xu, Bing, Warde-Farley, David, Ozair, Sherjil, Courville, Aaron, and Bengio, Yoshua. Generative adversarial nets. In NIPS, pp. 2672-2680, 2014.
	
	[4] Lake, B., Salakhutdinov, R., Gross, J., Tenenbaum, J. (2011). One shot learning of simple visual concepts. Proceedings of the Annual Meeting of the Cognitive Science Society, 33. Retrieved from
	
	[5] Karras, Tero, Aila, Timo, Laine, Samuli, and Lehtinen, Jaakko. Progressive growing of GANs for improved quality, stability, and variation. In ICLR, 2018.
	
	[6] Larsen, Anders Boesen Lindbo, S\o nderby, S\o ren Kaae, Larochelle, Hugo, and Winther, Ole. Autoencoding beyond pixels using a learned similarity metric. In ICML, pp. 1558-1566, 2016.
	
	[7] Makhzani, Alireza, Shlens, Jonathon, Jaitly, Navdeep, Goodfellow, Ian, and Frey, Brendan. Adversarial autoencoders. arXiv preprint arXiv:1511.05644, 2015.
	
	[8] Donahue, Jeff, Kr{\"a}henb{\"u}hl, Philipp, and Darrell, Trevor. Adversarial feature learning. In ICLR, 2017.
	
	[9] Danilo Rezende, Shakir, Ivo Danihelka, Karol Gregor, and Daan Wierstra. One-shot generalization in deep generative models. In Maria Florina Balcan and Kilian Q. Weinberger, editors, Proceedings of The 33rd International Conference on Machine Learning, volume 48 of Proceedings of Machine Learning Research, pages 1521-1529, New York, New York, USA, 20-22 Jun 2016. PMLR.
	
	[10] Sergey Bartunov and Dmitry P. Vetrov. Few-shot generative modelling with generative matching networks. In AISTATS, 2018.
	
	[11] Clou{\^a}tre, Louis, Demers M . FIGR: Few-shot Image Generation with Reptile[J]. 2019. 
	
	[12] Alex Nichol, Joshua Achiam, and John Schulman. On first order meta-learning algorithms, 2018.
	
	[13] Hochreiter, S. and Schmidhuber, J. Long short-term memory. Neural computation, 9(8):1735-1780, 1997.
	
	[14] Kaiming He, Xiangyu Zhang, Shaoqing Ren, and Jian Sun. Deep residual learning for image recognition. 2016 IEEE Conference on Computer Vision and Pattern Recognition (CVPR), Jun 2016.
	
	[15] Huang H, Li Z, He R, et al. IntroVAE: Introspective Variational Autoencoders for Photographic Image Synthesis[J]. 2018.
	
	[16] Jordan, Michael I, Ghahramani, Zoubin, Jaakkola, Tommi S, and Saul, Lawrence K. An introduction to variational methods for graphical models. Machine learning, 37(2):183-233, 1999.
	
	[17] Doersch C . Tutorial on Variational Autoencoders[J]. 2016.
	
	[18] Goodfellow, Ian, Bengio, Yoshua, Courville, Aaron, and Bengio, Yoshua. Deep learning, volume 1. MIT press Cambridge, 2016.
	
	[19] Lake, B. M., Salakhutdinov, R., Tenenbaum, J. B. (2015). Human-level concept learning through probabilistic program induction. Science, 350(6266), 1332-1338.
	
	[20] Ioffe S , Szegedy C . Batch normalization: accelerating deep network training by reducing internal covariate shift[C]// International Conference on International Conference on Machine Learning. JMLR.org, 2015.

\end{document}